\documentclass[10pt, letterpaper, conference]{IEEEtran}
\IEEEoverridecommandlockouts

\usepackage{color}

\usepackage{times}
\usepackage{epsfig}
\usepackage{graphicx}
\usepackage{tabularx}
\usepackage{amsmath}
\usepackage{amssymb}
\usepackage{bbm}
\usepackage[misc]{ifsym}
\usepackage[export]{adjustbox}
\usepackage{amsthm}
\usepackage{float}
\usepackage{graphicx} 

\usepackage{tikz}
\usepackage{verbatim}
\usepackage{booktabs}
\usepackage{multirow}
\usepackage{makecell}
\usepackage{authblk}
\usepackage{hyphenat} 

\usepackage{pgf}
\usetikzlibrary{arrows,automata}

\usepackage{graphbox}

\makeatletter
\newcommand{\removelatexerror}{\let\@latex@error\@gobble}
\makeatother

\makeatletter 
  \newcommand\figcaption{\def\@captype{figure}\caption} 
  \newcommand\tabcaption{\def\@captype{table}\caption} 
\makeatother

\makeatletter
\let\NAT@parse\undefined
\makeatother
\usepackage[colorlinks]{hyperref}  

\usepackage{graphics}
\usepackage{pifont}
\usepackage{times}

\usepackage{soul}
\usepackage{url}
\usepackage{graphicx}
\usepackage{amsmath}
\usepackage{booktabs}
\usepackage{balance} 
\usepackage[T1]{fontenc}    
\usepackage{hyperref}
\usepackage{wrapfig}
\usepackage{fancybox}
\usepackage{tikz}
\usepackage{algpseudocode}
\usepackage{algorithm}
\usepackage{diagbox}
\usepackage{multirow}
\usepackage{amsfonts}       
\usepackage{nicefrac}       

\usepackage{graphbox}

\usepackage[colorlinks]{hyperref}
\graphicspath{{figures/}}

\begin{document}
\title{{Communication-Efficient Reinforcement Learning in Swarm Robotic Networks for Maze Exploration}}
\author{Ehsan Latif$^1$ \and WenZhan Song$^2$ \and Ramviyas Parasuraman$^1$ 
\thanks{$^1$School of Computing, University of Georgia, Athens, GA 30602, USA \\
$^2$School of Electrical and Computer Engineering, University of Georgia, Athens, GA 30602, USA \\
Corresponding author email: ramviyas@uga.edu}}

\maketitle              

\begin{abstract}
Smooth coordination within a swarm robotic system is essential for the effective execution of collective robot missions. Having efficient communication is key to the successful coordination of swarm robots. This paper proposes a new communication-efficient decentralized cooperative reinforcement learning algorithm for coordinating swarm robots. It is made efficient by hierarchically building on the use of local information exchanges. We consider a case study application of maze solving through cooperation among a group of robots, where the time and costs are minimized while avoiding inter-robot collisions and path overlaps during exploration. With a solid theoretical basis, we extensively analyze the algorithm with realistic CORE network simulations and evaluate it against state-of-the-art solutions in terms of maze coverage percentage and efficiency under communication-degraded environments. The results demonstrate significantly higher coverage accuracy and efficiency while reducing costs and overlaps even in high packet loss and low communication range scenarios.

\end{abstract}
\begin{IEEEkeywords}
Swarm, Robotic Networks, Path Planning, Reinforcement Learning, Maze Exploration, Communication
\end{IEEEkeywords}

\section{Introduction}
\label{sec:intro}

Maze coverage and exploration-related challenges have always piqued the interest of humans. 
Path planning algorithms are typically used to regulate how robots move through mazes if the structure of the labyrinth (including its walls and paths) is already known. If the maze structure is unknown, the robots must first use sensors to find a portion of the maze nearby before planning their next move based on the knowledge they have gained. Exploring an unknown maze in pursuit of stationary targets is thus a challenge associated with coverage, search, and path-finding issues \cite{alamri2021autonomous}.
Typical maze coverage solutions such as A* \cite{tjiharjadi2017optimization}, and Depth First Search \cite{chen2020improved} apply to a single robot system and require constant updates and optimization. However, most current research focuses on coordinated multi-robot and swarm robotic networks \cite{youssefi2021swarm}.

Numerous research has solved the issue of coordinating and controlling several robots for mapping and exploration. However, most strategies rely on centralized control to guide each robot in a swarm because centralized coordination enables almost optimal behaviors in surroundings that are well understood. However, in the novel, unknown contexts, a distributed approach for swarm coordination is necessary, as was put forth by Bono et al. \cite{bono2021swarm}, where it produced a reliable result even if one or more of the robots were lost.
Specifically, sharing locally observed maze information among the connected robots is a crucial component of distributed approaches so that each agent can carry out its plan without interference from the others. However, a reliable information fusion technique needs to be integrated with such approaches for communication efficiency and a reliable system for information fusion \cite{kalinowska2022over}. 

\begin{figure}[t]
\centering
\begin{center}
 \includegraphics[width=0.98\columnwidth]{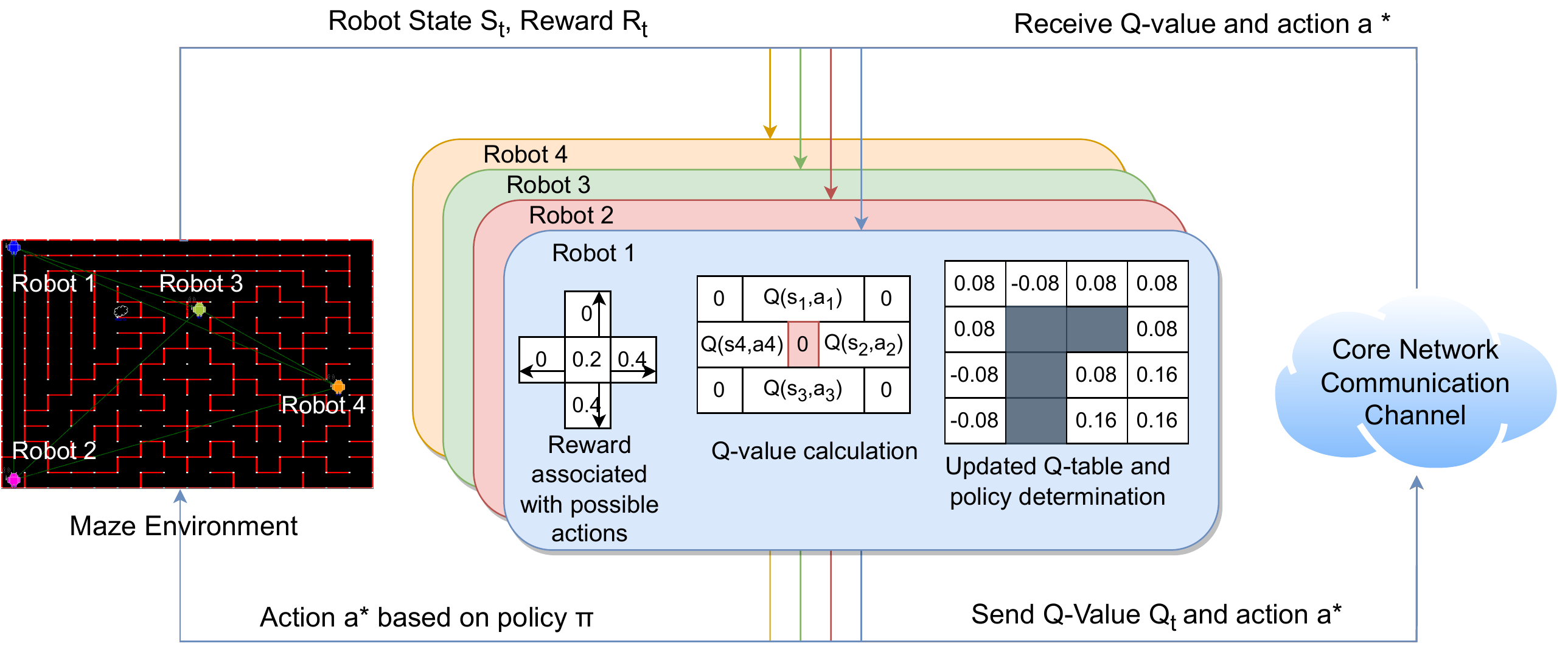}
\end{center}
\vspace{-2mm}
 \caption{Overview of the Communication-Efficient Reinforcement Learning.}
 \label{fig:RL}
 \vspace{-4mm}
\end{figure}

Reinforcement learning (RL) is a machine learning technique that trains agents to take action in an environment to maximize a reward signal. In swarm robotic networks, the robots can be trained to explore the maze effectively by learning from their interactions with the environment. This problem is challenging because the robots may have different capabilities, and the maze is unknown.

Multi-robot maze coverage using RL has been recently explored for solving the problem of coordinating a team of robots to explore and map a maze \cite{gu2021improved}. 
However, recent RL algorithms fail to scale up quickly for a swarm system and do not work efficiently in communication-constrained environments where the packet loss rates are high, and the communication range may be limited. 
Therefore, to remedy this issue, we contribute a new efficient novelty search through deep reinforcement learning (DRL) \cite{shi2020efficient} to coordinate a group of robots and solve the maze exploration problem distributedly; an overview of the proposed approach can be seen in Fig.~\ref{fig:RL}.

One approach to solving the swarm robotic maze exploration using RL is to define the problem as a Markov Decision Process (MDP), where the states represent the positions of the robots and the walls of the maze, the actions represent the movements of the robots, and the rewards represent the progress made towards exploring the maze. The robots can then use an RL algorithm, such as Q-learning, to learn a policy that maps states to actions and maximizes the rewards in a distributed manner.

We theoretically analyze our approach with other methods from the literature. We extensively evaluated the proposed algorithm using CORE network simulations in different maze worlds. This paper focuses on the efficiency of the proposed approach for micromouse maze coverage under standard networking conditions; hence we are not discussing networking aspects of the work. Comparing the results from a graph-based Depth First Search (DFS) \cite{chen2020improved} and the recent memory-greedy reinforcement learning \cite{yu2021memory} approaches, the proposed RL algorithm outperforms in terms of coverage efficiency and minimizing overlapping regions, especially in communication-degraded environments.
We open-source the codes in GitHub at \url{https://github.com/herolab-uga/MazeCommRL} for use and further development by the robotics community.
The algorithm can be generalized to other application domains where communication efficiency is crucial for better swarm coordination.

\section{Related Work}
\label{sec:relatedwork}

In scenarios where both the world (map of the maze) is unknown and multiple robots need to coordinate effectively to explore the full map, the robots must search the uncharted maze to locate a target. 
Many algorithms have been proposed in the literature. For instance, a search-based method to locate a path out of the maze may not always provide the shortest path \cite{tjiharjadi2022systematic}. 
It is also becoming increasingly important to plan how to address the multi-robot pathfinding that more effectively bypasses cost estimations during preparation \cite{luis2020using}. In \cite{shi2019multi}, an optimization-based UAV trajectory planning method is proposed to enable a drone-assisted open radio access network. 
In \cite{wang2019coordination}, a language-based explicit communication is introduced for swarm robots to coordinate and plan their tasks with low communication costs.

However, most of them assume a centralized control or coordination system. The robots do not work together effectively, making it impossible for a swarm robotic network to perform well without significant memory and communication overload.

Reinforcement learning provides a superior way to tackle communication-efficient swarm coordination challenges. This is because robots' actions impact other robots' actions and vice versa, which encourages robots to collaborate by utilizing other robots' information through communication \cite{pozza2021quantum,yang2021can}. 
Such information benefits the robots' cooperation, but they cannot ensure they will communicate all the information they need to exchange to collaborate.

In the maze problem, the robot will receive a positive reward when it arrives at its destination, meaning that the reward it just received will allow it to return to the original region numerous times. These prizes, however, only show up one time in the maze. This circumstance, known as a "reward loop," significantly extends the time needed to navigate the maze. 
A Q-learning method with multiple Q tables is suggested in a study \cite{kantasewi2019multi}. This approach reduces map area overlap between robots by creating a Q table each time the robot explores a new region, allowing the robot to navigate to the target optimally.
The Q-learning is expanded in \cite{uwano2017communication}, and an RL for multi-robot cooperative tasks is proposed. The standard reward is transformed into an "internal reward" when robots in the maze cannot communicate with one another. This allows the robots to learn following this "internal reward" type and achieve task collaboration under such challenging circumstances.

A unique RL maze navigation technique is proposed in \cite{yu2019navigation} to address the labyrinth navigation problem for robotic vehicles. First, the picture data of the random maze is collected using a drone's bottom-mounted camera. Then the virtual maze is created in the simulation environment using an image processing technique. An enhanced Q-learning algorithm is suggested to address the issue that the original greedy strategy repeatedly forces the robot to linger in the past state.
However, it cannot tolerate significant communication degradation.
Alternatively, our proposed approach considerably differs from the literature by considering an efficient information transfer mechanism combined with Q-learning for multi-robot cooperation to solve maze exploration tasks.


\section{Proposed Approach}
\label{sec:problem}
Let us assume multiple robots of different types are tasked with exploring and mapping a maze. Each robot has unique capabilities and strengths, such as different speeds and sensor ranges. Using reinforcement learning, the robots learn to cooperatively navigate the maze and cover as much as possible. The robots communicate and coordinate with each other to efficiently cover the maze and avoid collisions. We assume the robots can localize in a global frame of reference (e.g., GPS); however, recent relative localization techniques like \cite{latif2022dgorl} can be employed if global positioning is unavailable.

As the robots explore the maze, they receive rewards for discovering new areas and penalties for collisions. Over time, the robots learn the optimal strategies for exploring the maze and maximize their coverage. Each robot decides its action plan based on the below approach.

\subsection{Reinforcement Learning}
Popular machine learning algorithms like reinforcement learning use interactions with the environment to teach robots new skills. Markov decision processes are frequently used to model it. The robot's state is one of them and is denoted by the letter $s$. A state transition from state $s_t$ to state $s_{t+1}$ occurs due to the reinforcement learning robot selecting and carrying out an action based on its state at time $t$. The robot will receive a reward for each action. The robot will have learned the course of action to take in each condition and will be able to maximize the reward of the entire process through repetitive trial and error. Fig.~\ref{fig:RL} depicts the fundamental idea of RL, and the symbols and concepts used in RL are as follows:
\begin{itemize}
    \item $S = \{s_1, s_2, s_3,...,s_n\}$ is a discrete set of $n$ states, where $s_t \in S$ describes the state of the robot in the environment at a time $t$.

    \item $A = \{a_1, a_2, a_3,..., a_n\}$ is a discrete set of $n$ actions, where $a_t \in A$ describes the action which the robot chooses at a time $t$.

    \item $T : S \times A \times S \rightarrow [0, 1]$ is a stochastic state transition function, where the state of the robot is transitioned to state $s^*$ with a probability $p \in [0, 1]$ when choosing action $a$ in state $s$. We use $s^* \leftarrow T (s, a)$ to represent the above process.

    \item $R: S \times A \times S \rightarrow R$ is the reward function. It represents the robot's reward in its state transition to $s^*$ after executing action $a$ in state $s$.

    \item $\gamma \in [0, 1)$ is the discount factor is the relative importance of future and present rewards.
\end{itemize}

Kantasewi et al., \cite{kantasewi2019multi} suggest a Q-learning with a multi-table approach. It is a technique that can be continuously updated in light of prior knowledge to ultimately arrive at the most likely accurate choice. The Q table is the Q learning's secret. All possible states and actions are created using the Q table, which then updates each value through iterative learning. The robot then chooses the best course of action for each state based on the values in the table. This approach is frequently utilized in path planning, chess, card games, and other activities.
The objective of the proposed approach is to perform maximum coverage in less time and avoid overlapping exploration, which can be numerically defined as:
\begin{equation}
    \max_\pi \{P_a^\pi (t) - \lambda E_t(a|\pi)\} ,
\end{equation}  
where $P_a^\pi(t)$ is the probability of covering the region for action $a$ using policy $\pi$ at time $t$,  $E_t(a|\pi)$ is an average number of steps to cover associated with action an $a$ at time $t$ in policy $\pi$ and $\lambda$ is the cost associated with each step.

The algorithm updates the Q value by the following formula:
\begin{align*}
& {Q_{t + 1}}\left( {{s_t},{a_t}} \right) \\
& = (1 - \alpha ){Q_t}\left( {{s_t},{a_t}} \right) + \alpha \left[ {{r_t} + \gamma \max {Q_t}\left( {{s_{t + 1}},{a_t}} \right)} \right] , \tag{1}
\label{eqn: q-value}
\end{align*}
where $s_t$ and $s_{t+1}$ are the current states and the next state, respectively, it is the action executed, $\alpha \in (0, 1]$ controls the balance between the coverage and delay, and $\gamma$ is the discount factor.
 At each discrete time step $t$, the agent acquires an observation $s_t$ from the environment, selects a corresponding action $a_t$, then receives feedback from the environment in the form of a reward $r_t=R(s_t,a_t)$ as:
\begin{equation}
R_{t} = \sum_{i=1}^{N} \left[ \alpha_i \cdot \left( \frac{1}{d_{i,t}} - \frac{1}{d_{i,t-1}} \right) + \beta_i \cdot \left( 1 - \frac{1}{d_{i,t}} \right) \right] ,
\end{equation}
where $N$ is the number of robots, $d_{i,t}$ is the distance of a robot $i$ from the goal at time $t$, $\alpha_i$ and $\beta_i$ are constants that represent the reward for each robot.
 
 The updated state information $s_{t+1}$. The goal of the RL agent is to select policy $\pi$ to maximize the discounted sum of future rewards, i.e., $Q_\pi(s_1) = \sum_\tau^ {t=1}\gamma^t R(s_t,a_t)$, which according to the Bellman optimality principle satisfies.

 This   reward   function   produces   a   negative reward  whenever  the  agent  has  looped  back  and  no  reward otherwise.
Alg.~\ref{alg:HMMC_reinforcement_learning} presents the pseudocode description of the approach using reinforcement learning.

\subsection{Swarm robot cooperation}
The next issue we have addressed is the communication amongst individual exploration-capable robots in our multi-robot system. Our proposed RL-based exploration system is designed for a single robot, allowing each robot to make independent decisions based on local information and with little interaction from other robots. We introduce efficient communication amongst nearby robots to encourage cooperation. We develop a discovery approach based on the distance between simulated robots to replicate the network range in which we only share the current position of a $robot_i$, its Q-Value for each direction, and mark the current situation as explored to avoid repetitive exploration. When another robot receives the information, it will update the received Q value in its Q table and update the local map.

\begin{table*}[ht]
\begin{center}
\caption{Comparative complexity analysis of maze exploration.}
\label{table2}
\vspace{-4mm}
\resizebox{\textwidth}{!}{
\begin{tabular}{|c|c|c|c|c|c|c|c|}
\hline
\textbf{Algorithm}            & \textbf{Category}                                            & \textbf{Best Case}     & \textbf{Worst Case}    & \textbf{Average Case}
 & \textbf{Coverage} & \textbf{Update Cost}  & \textbf{Abbreviations}  \\ \hline
 
Ant colony optimization \cite{viseras2016planning} & Centralized  & $O(n \times m)$ & $O(2 n \times m)$ & $O(b\times  n\times m)$ & partial & low & b - no. of obstacles, m-no. of robots, n - no. of cells                    \\
\hline
Cooperative Random Walk \cite{ozdemir2019spatial} & Decentralized  & $O(log n)$ & $O(n^3)$ & $O(n^{2/3})$ & partial & low &  n - number of cells                    \\
\hline
Improved SAT \cite{surynek2016efficient} & Centralized  & $O(\mu \times n\times m)$ & $O(\mu \times n\times m)$ & - & full & high & $\mu$ - makespan factor                   \\
\hline
Efficient DRL \cite{shi2020efficient} & Decentralized  & $O(\epsilon \times n)$ & $O(\epsilon \times n)$ & - & full & high &  $\epsilon$ - depth factor                   \\
\hline
Memory-Greedy RL \cite{yu2021memory} & Decentralized  & $O(\epsilon_2 \times n)$ & $O(\epsilon_2 \times n)$ & - & full & high &  $\epsilon_2$ - greedy degree                 \\
\hline
Proposed RL Approach & Decentralized  & $O(k)$ & $O(k)$ & - & full & moderate &  k - number of sub-mazes                 \\
\hline
\end{tabular}
}
\end{center}
\vspace{-3mm}
\end{table*}

\begin{algorithm}[t]
\caption{Communication-Efficient RL for Exploration}
\label{alg:HMMC_reinforcement_learning}
\begin{algorithmic}[1]
\State Initialize the maze environment and the agent's state $s_0$
\State Initialize the reward functions $R_1, R_2, \dots, R_n$
\For{each episode}
    \State Initialize the episode's total reward $R = 0$
    \For{each step in the episode}
        \State Take action $a$ according to the agent's policy $\pi$
        \State Observe the reward $r$ from the environment
        \State Update the total reward $R = R + r$
        \State Update the agent's state $s_{t+1} = f(s_t, a)$
    \EndFor
    \For{each reward function $R_i$}
        \State Calculate the reward $r_i = R_i(s_0, s_T)$
        \State Update the total reward $R = R + r_i$
    \EndFor
\State Update agent's policy $\pi$ using gradient ascent on $R$ 
\EndFor
\end{algorithmic}
\end{algorithm}

\subsection{Procedure}
Based on the RL and cooperation strategy discussed above, our approach incorporates cooperation into Q learning. A multi-robot cooperation strategy learning method based on RL has been suggested to solve the issue that the state action space is too large due to the complex maze environment when using traditional Q-learning to solve maze problems, resulting in too many iterations and too long in the entire learning process. These are the precise stages:
\begin{itemize}
    \item Given the target maze $M$, it is divided into multiple sub-mazes. We regard these sub mazes as a set $m$, where $m = {m_1, m_2, m_3,..., m_n}$.

    \item Multiple robots use Q-learning to explore these sub-mazes at the same time. Because the sub-maze is much simpler than the target maze, it only needs a few iterations to complete the learning.

    \item Each robot updates its $Q$ table at every iteration based on Q-value update Eq.~\eqref{eqn: q-value}. After learning, we can get a set of $Q$ tables, where $Q = \{Q_1, Q_2, Q_3,..., Q_n\}$.

    \item Combine these $Q$ tables as the initialized $Q$ table of the target maze, called $Q_{cop}$.

    \item Finally, based on $Q_{cop}$, the robot completes the exploration of the target maze $M$.
\end{itemize}

\subsection{Theoretical Analysis}

We have compared our proposed approach with state-of-the-art algorithms from four major categories for maze exploration: Ant colony optimization \cite{viseras2016planning} (search-based); cooperative random walk \cite{ozdemir2019spatial} (randomized); Efficient SAT \cite{surynek2016efficient} (reduction-based); and Efficient DRL \cite{shi2020efficient} (machine learning). Ant colony optimization is a centralized algorithm that provides partial coverage with a complexity of $O(2n\times m)$ for $n$ robots in a maze of $m$ cells. Cooperative random walk operates purely randomly and cannot guarantee full maze coverage with a high time complexity of $(n^3)$ for worst-case scenarios. Improved SAT using a lower bound on the sum of costs and an upper bound on the makespan provides complete coverage of the maze with high update cost. Efficient DRL is the most comparative approach and has $\epsilon$ depth factor for exploration and shares the whole Q table in every iteration but can operate efficiently to cover the entire map in less time with full cooperation. Our proposed approach works in a decentralized fashion; each robot calculates the Q value for its current cell, shares only the current value, and updates the Q table accordingly; hence, our approach has low communication/update cost and only traverses a single time through each cell to provide full maze coverage. 
Table~\ref{table2} depicts that the proposed approach provides full maze coverage with reasonable computational efficiency and low update cost.

\begin{figure}[t]
    \centering
\includegraphics[width=0.39\linewidth]{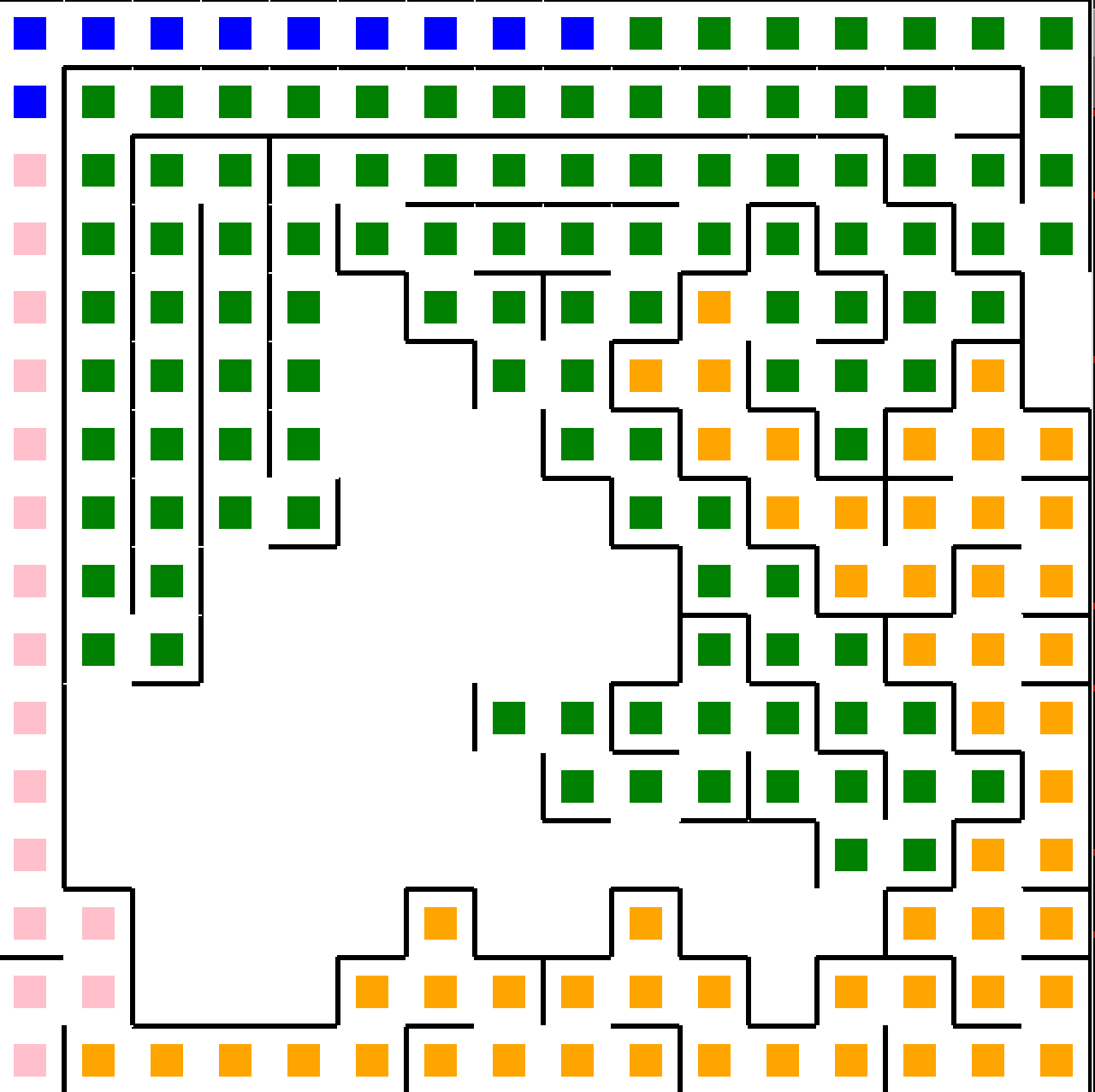}
\includegraphics[width=0.58\linewidth]{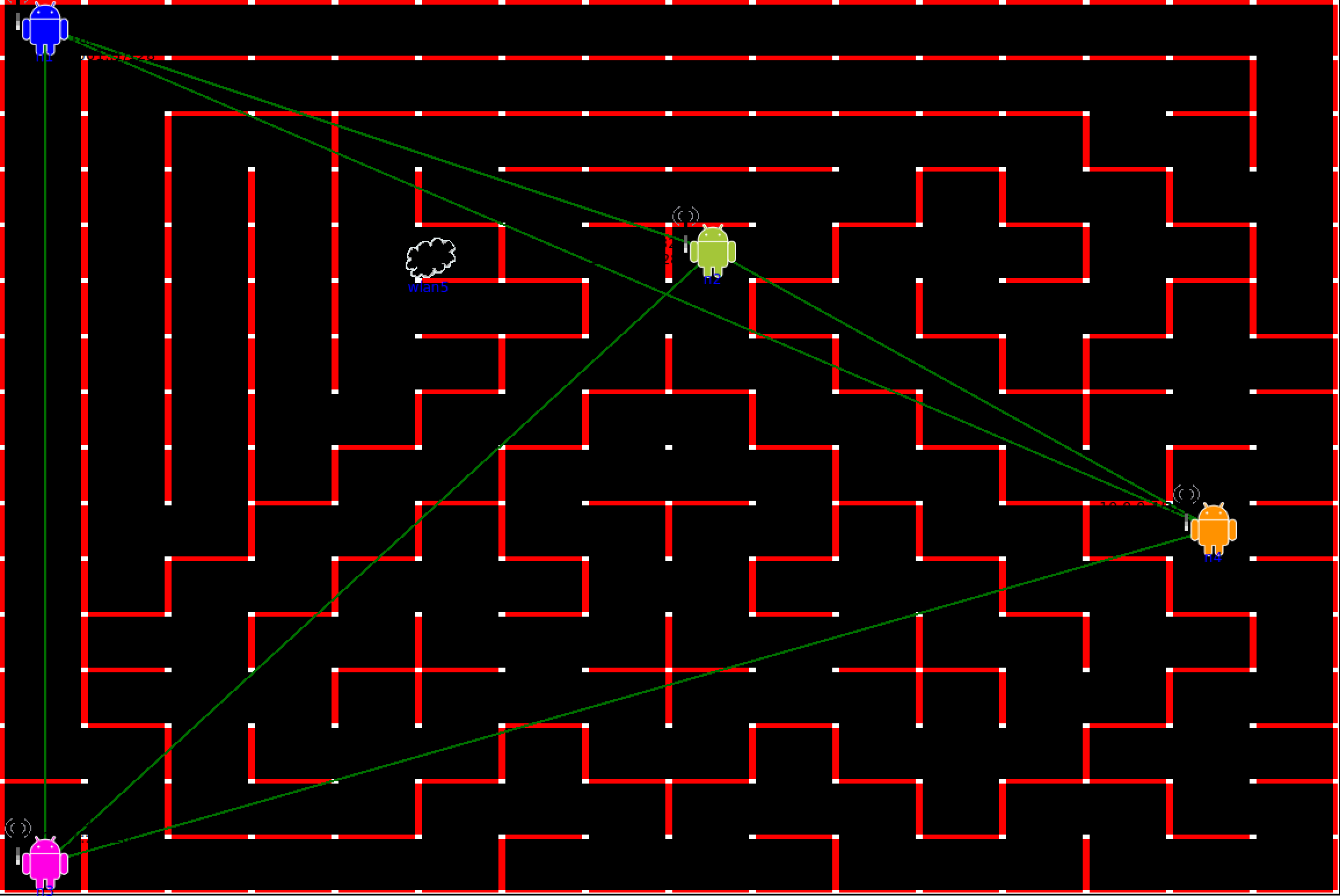}
\caption{Core Simulator \textbf{(Right)} with exploration map \textbf{(Left)}.}
    \label{simulation}
    \vspace{-6mm}
\end{figure}

\begin{figure}[t]
\centering
\begin{center}
 \includegraphics[width=0.9\columnwidth]{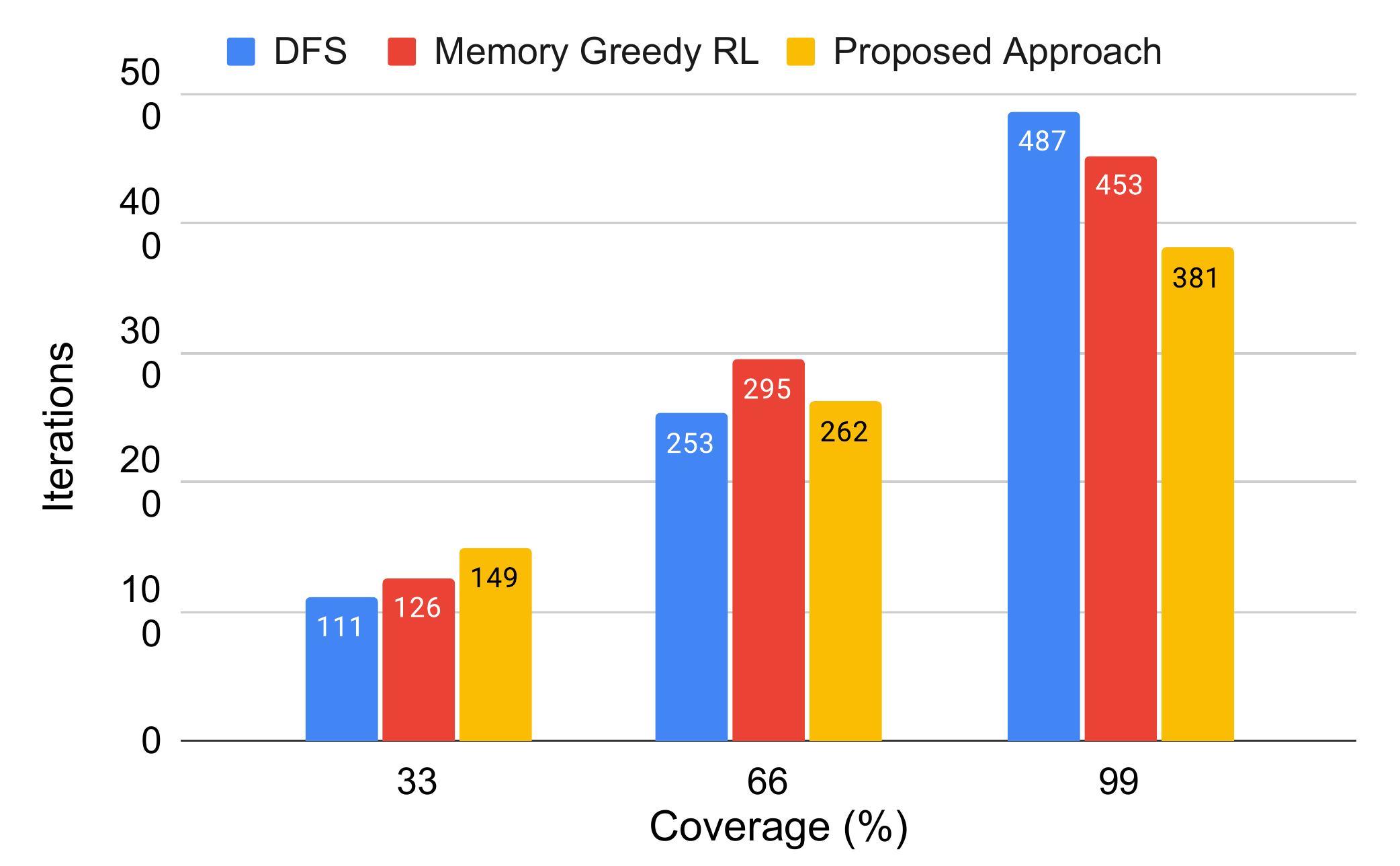}
\end{center}
\vspace{-4mm}

 \caption{Time taken by maze coverage strategies for three levels of coverage (one-third, two-third, and full).}
 \label{time_per_coverage}
 \vspace{-6mm}
\end{figure}

\begin{figure*}[ht]
    \centering
\includegraphics[width=0.32\linewidth]{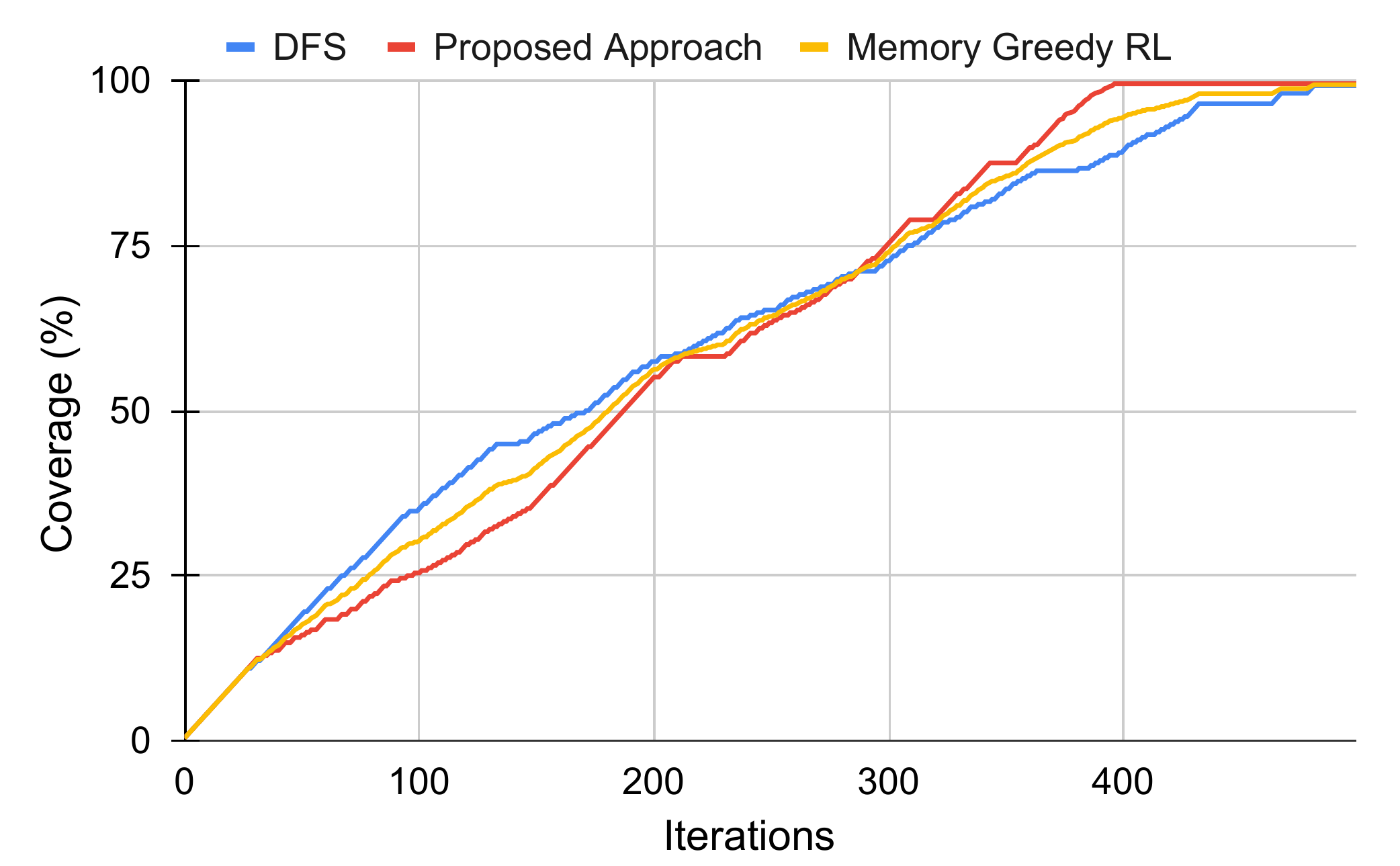}
\includegraphics[width=0.32\linewidth]{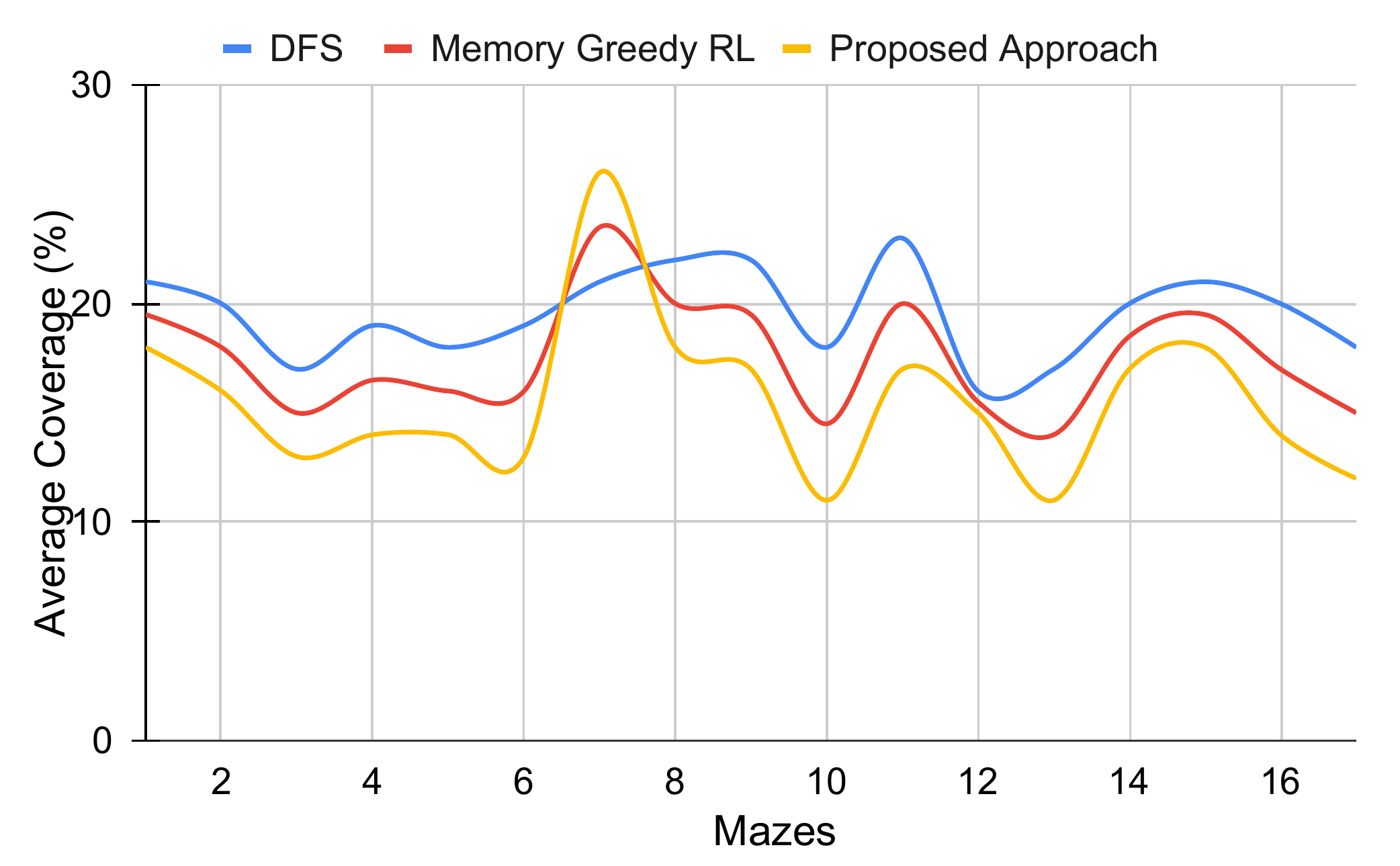}
\includegraphics[width=0.32\linewidth]{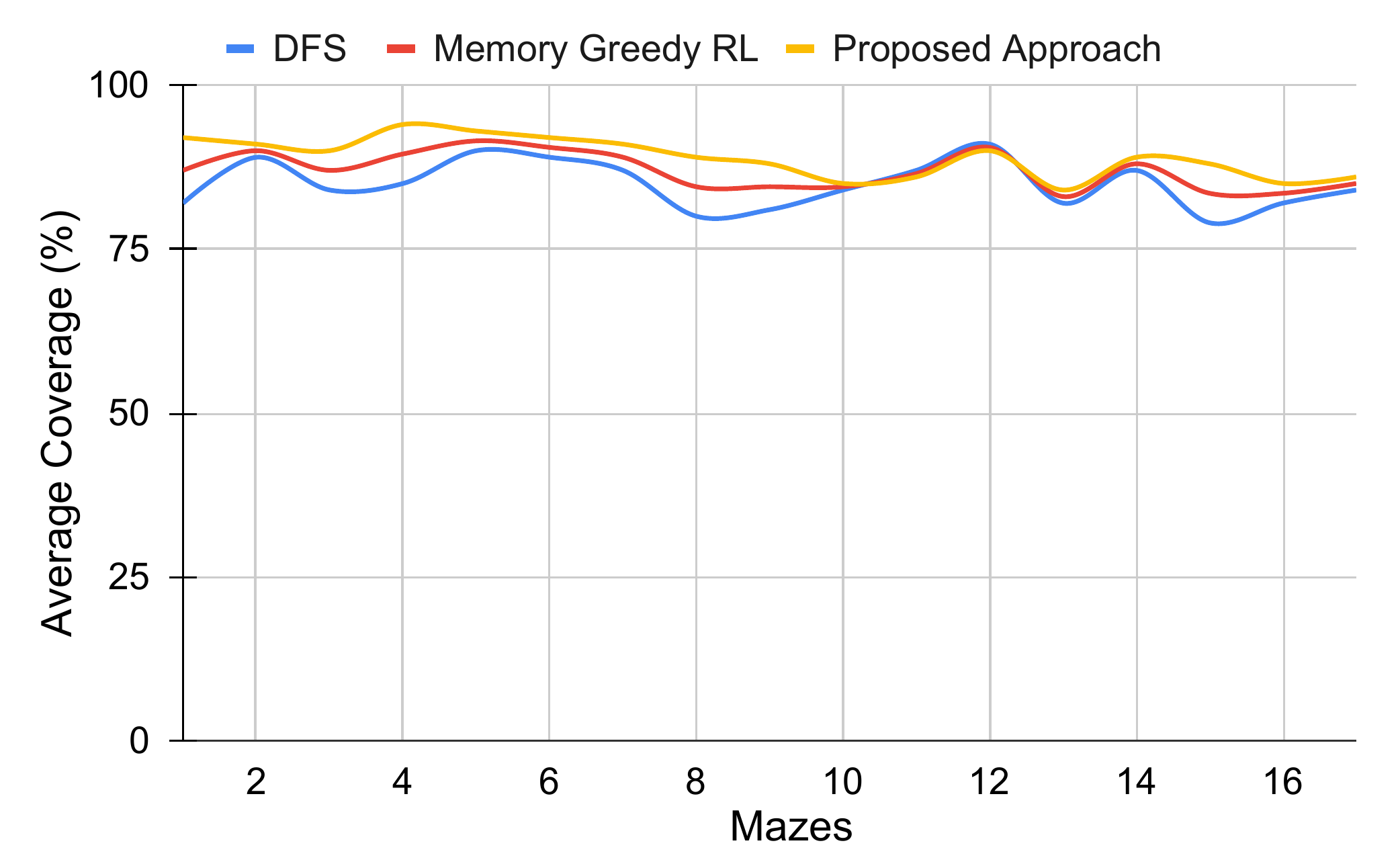}
\vspace{-4mm}
\caption{Coverage percentage performance metric: Average results over all mazes \textbf{(Left)}, after 100 iterations \textbf{(Center)}; after 400 iterations \textbf{(Right)}.}
    \label{coverage_results}
\end{figure*}

\begin{figure*}[ht]
    \centering
\includegraphics[width=0.32\linewidth]{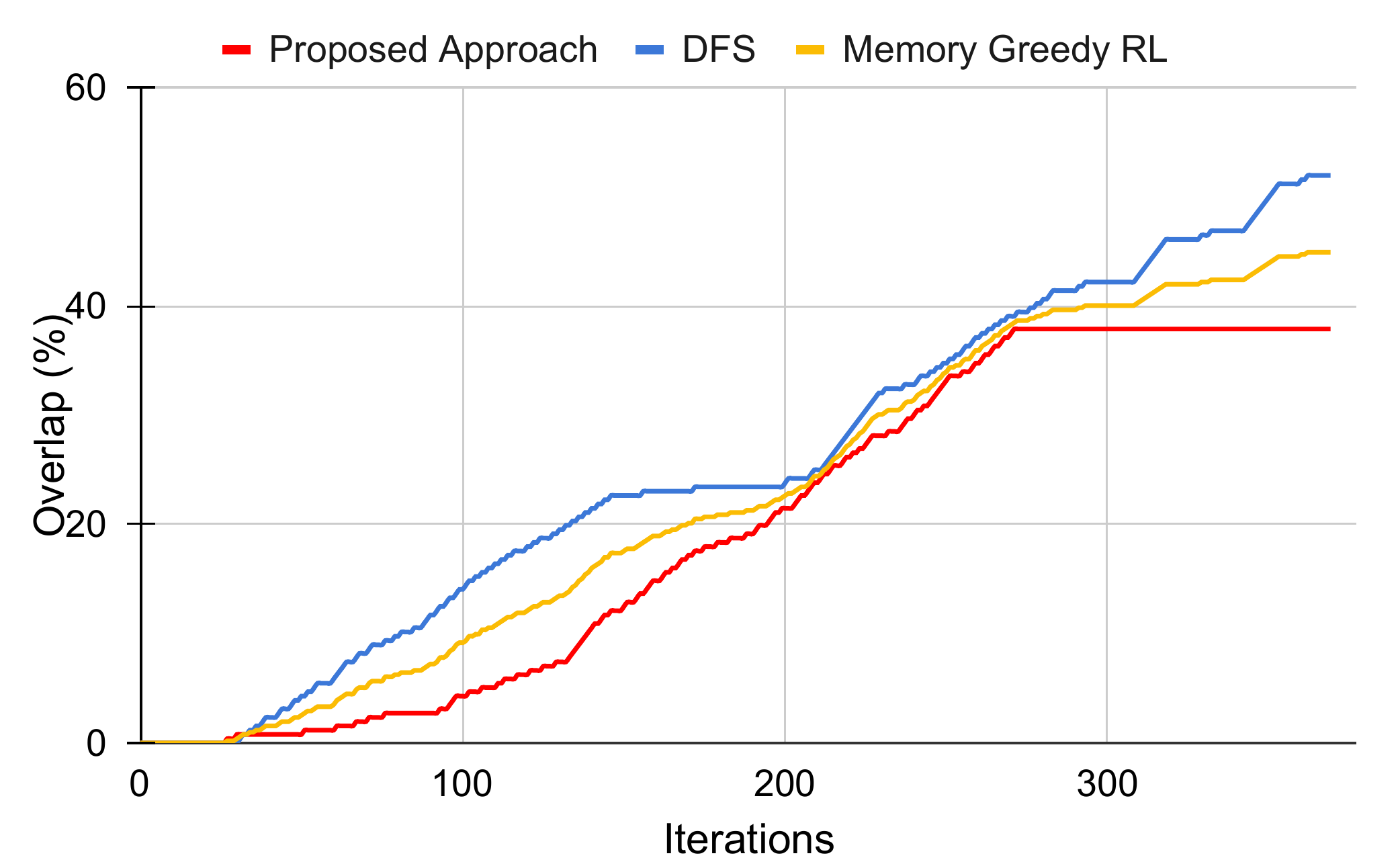}
\includegraphics[width=0.32\linewidth]{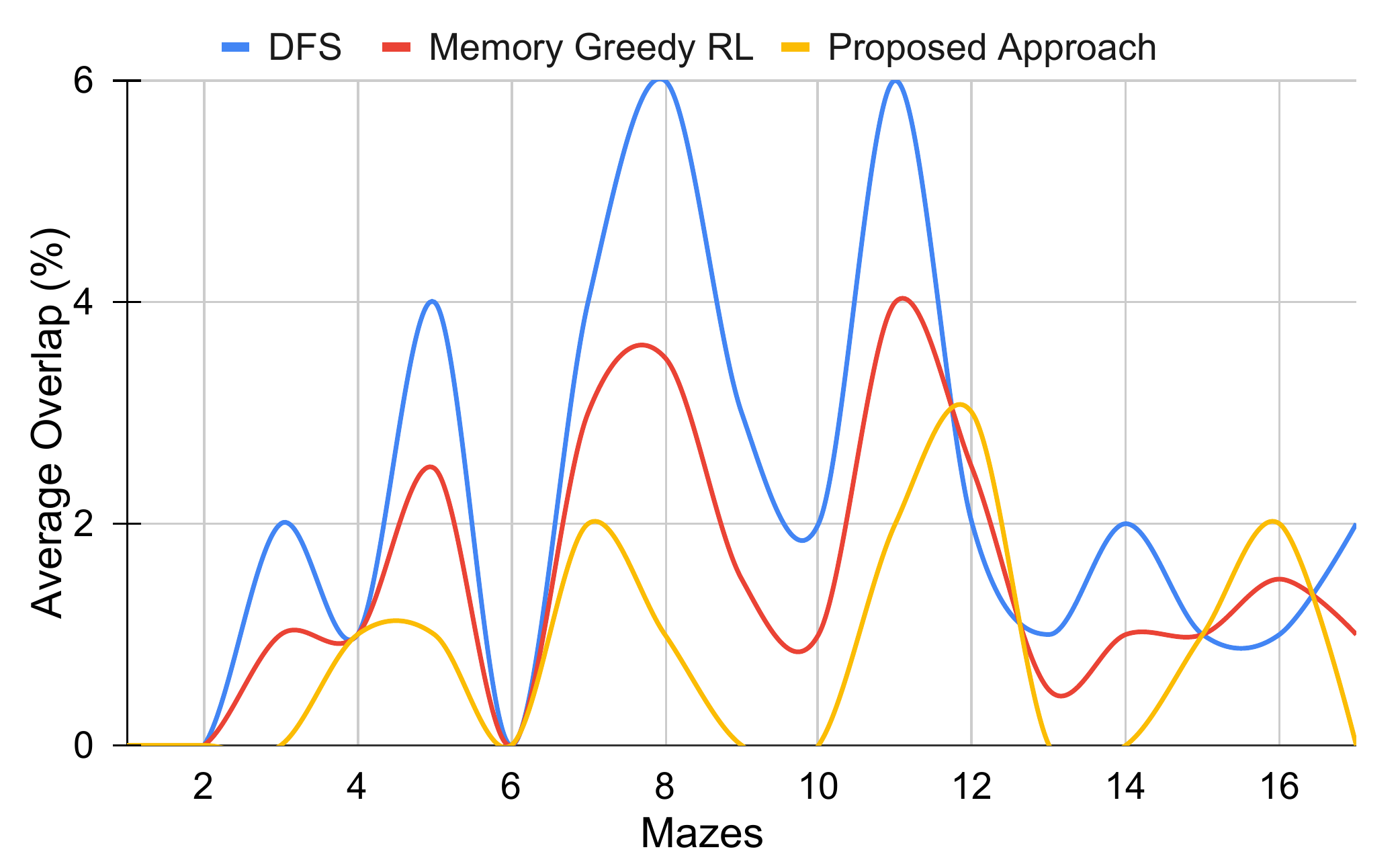}
\includegraphics[width=0.32\linewidth]{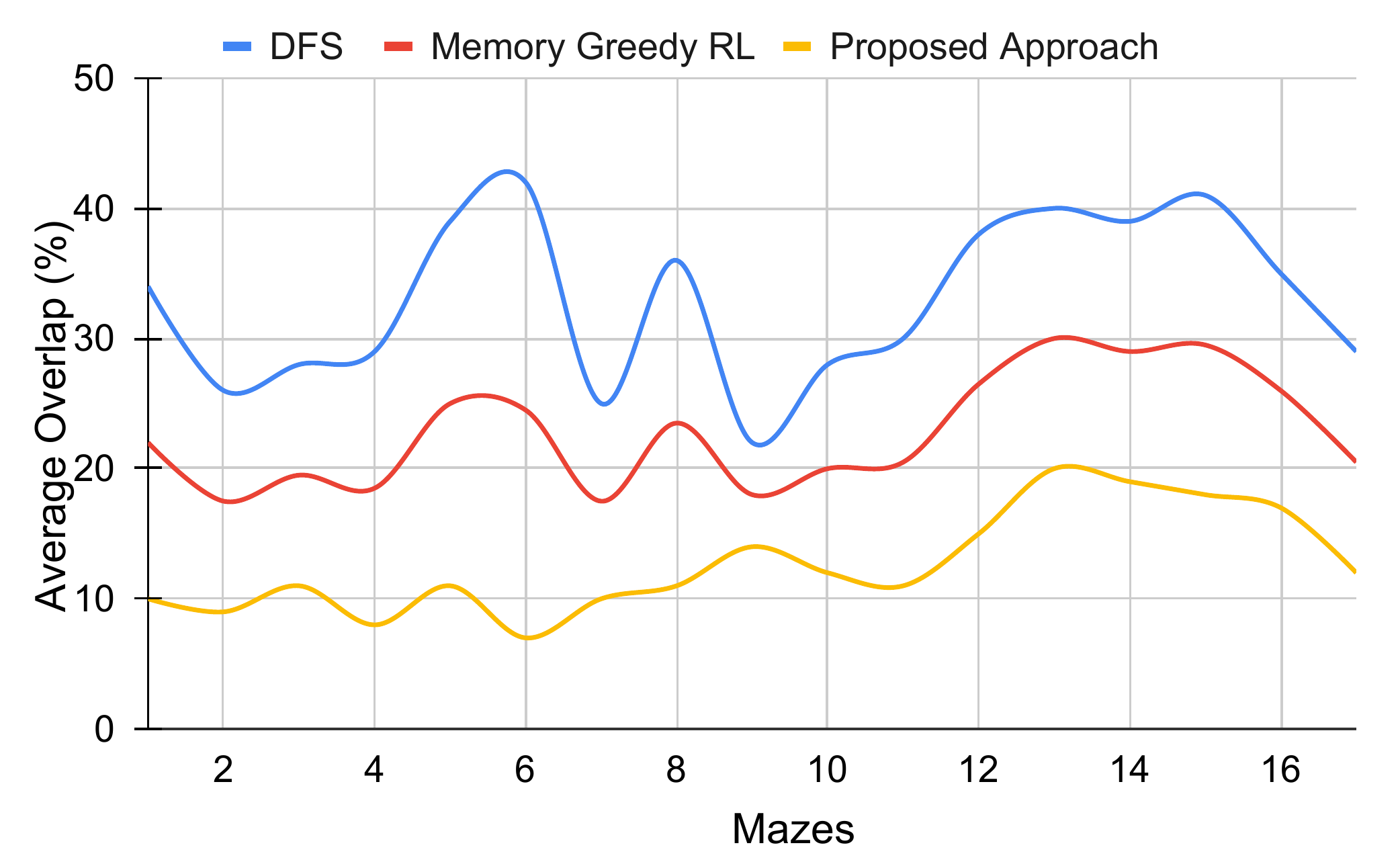}
\vspace{-4mm}
\caption{Map overlap percentage metric: Average results over all mazes \textbf{(Left)}, after 100 iterations \textbf{(Center)}; after 400 iterations \textbf{(Right)}.}
    \label{overlap_results}
\end{figure*}

\begin{figure*}[ht]
    \centering
\includegraphics[width=0.31\linewidth]{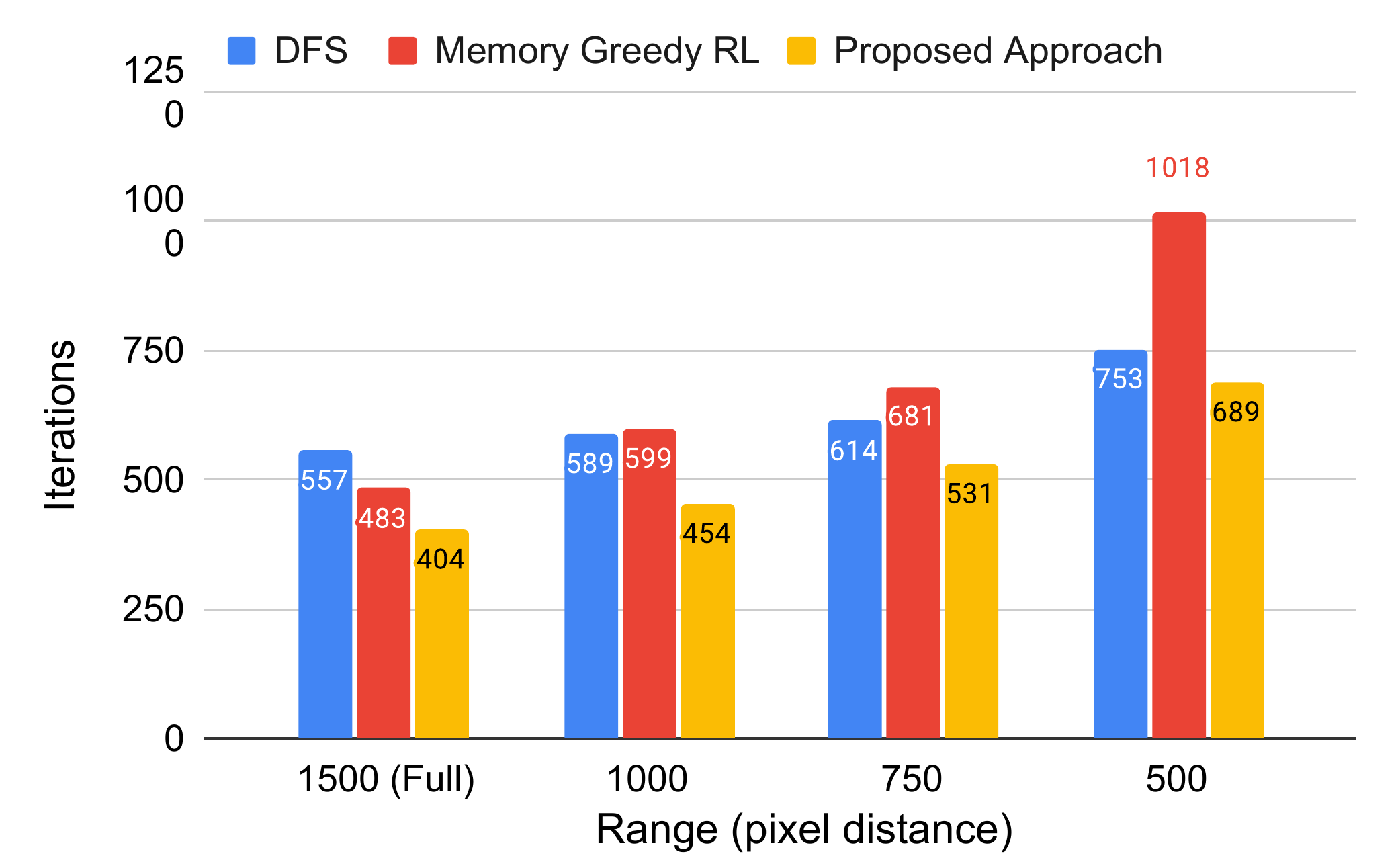}
\includegraphics[width=0.31\linewidth]{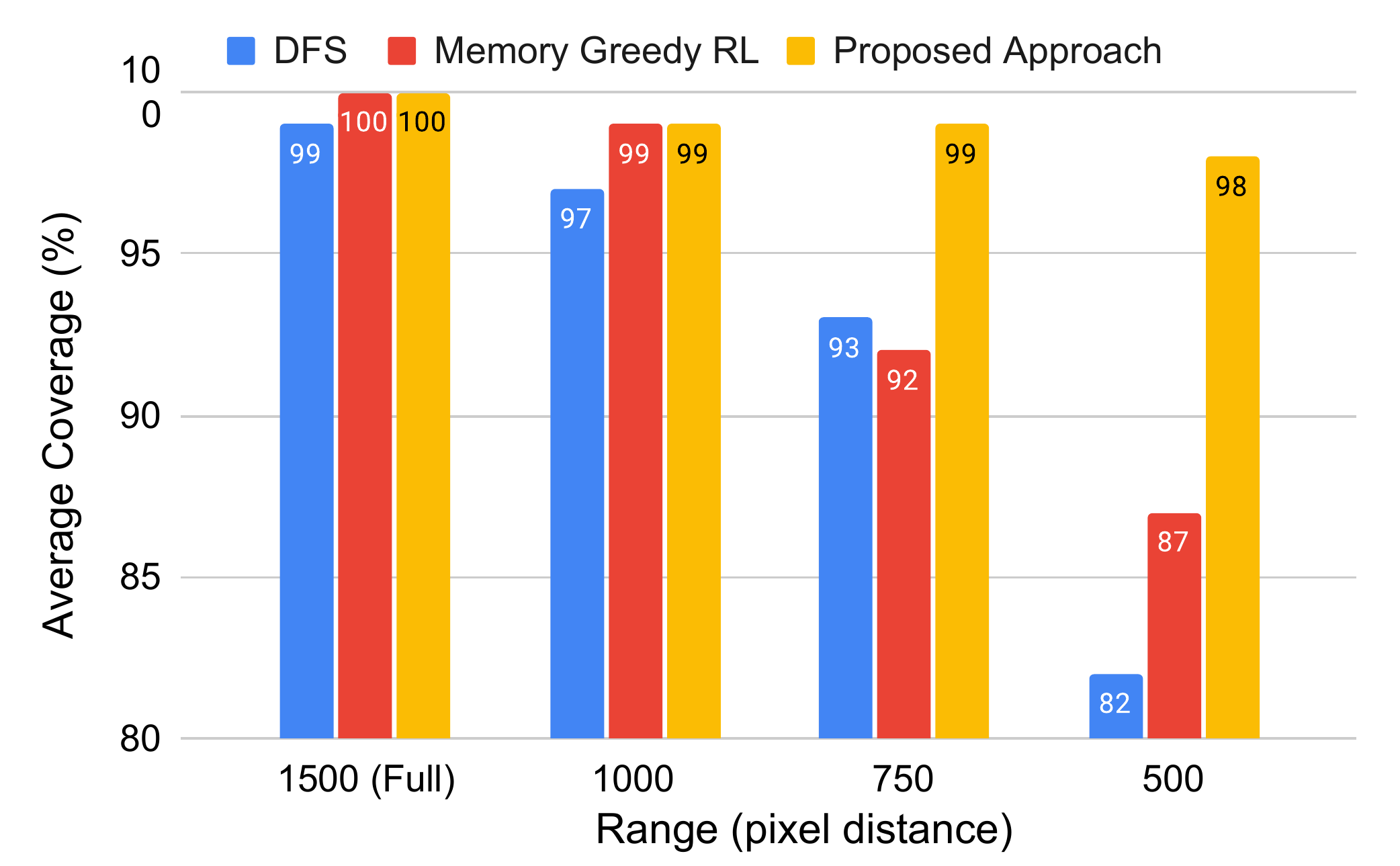}
\includegraphics[width=0.31\linewidth]{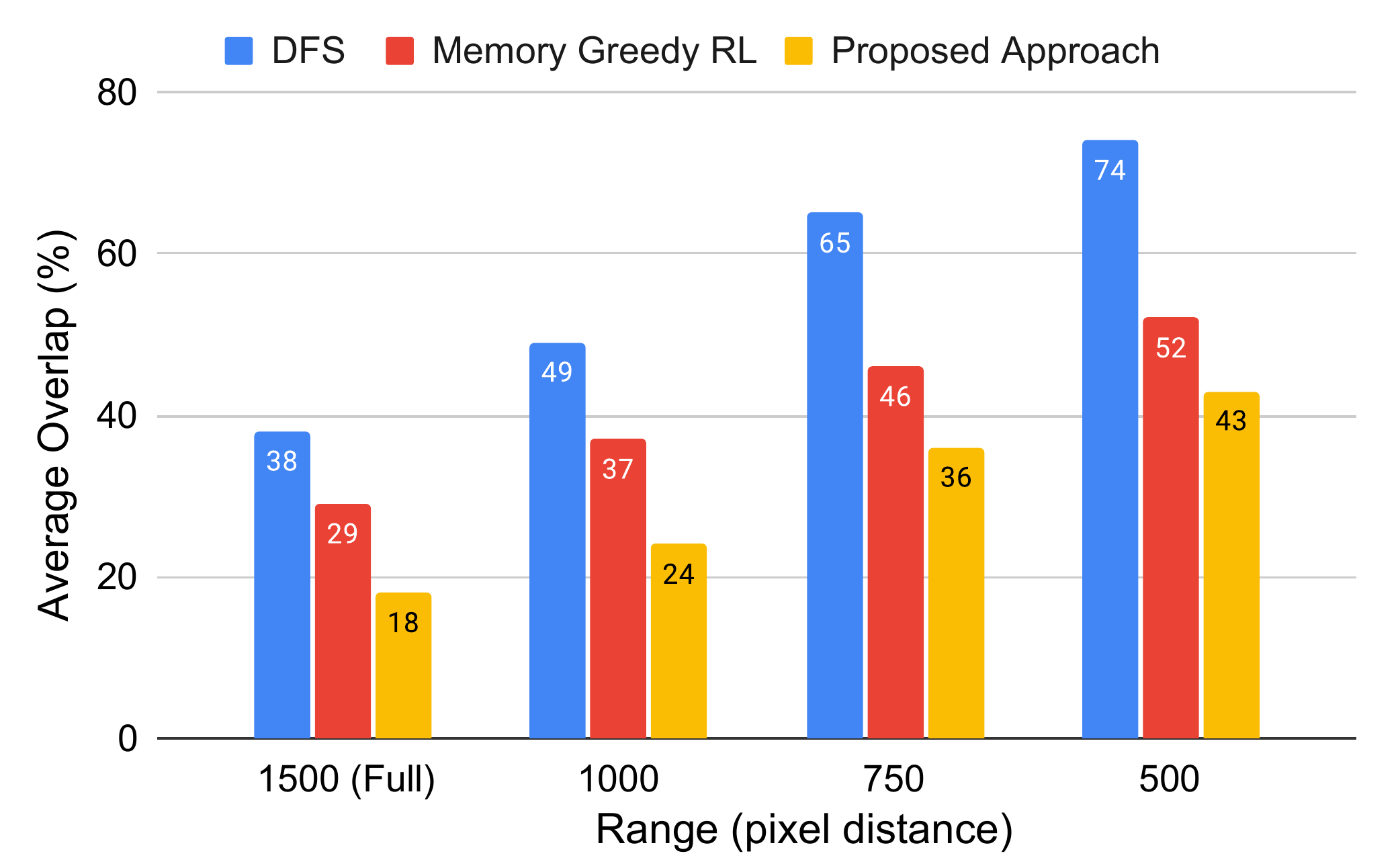}
\vspace{-4mm}
\caption{Impact of the communication range parameter; number of iterations \textbf{(Left)}, average coverage percentage \textbf{(Center)}, average overlap percentage \textbf{(Right)}.}
    \label{range_results}
\end{figure*}

\begin{figure*}[ht]
    \centering
\includegraphics[width=0.31\linewidth]{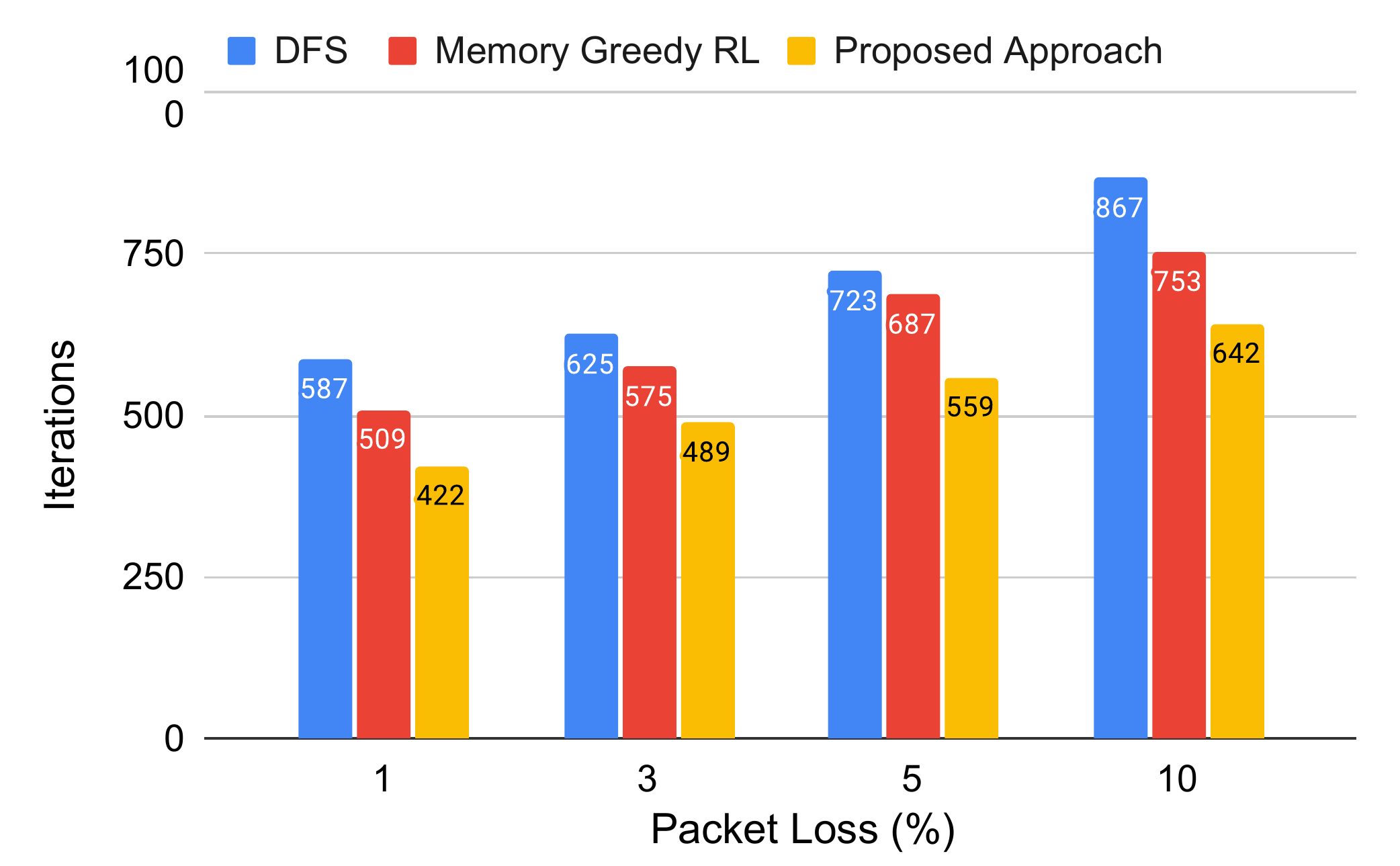}
\includegraphics[width=0.31\linewidth]{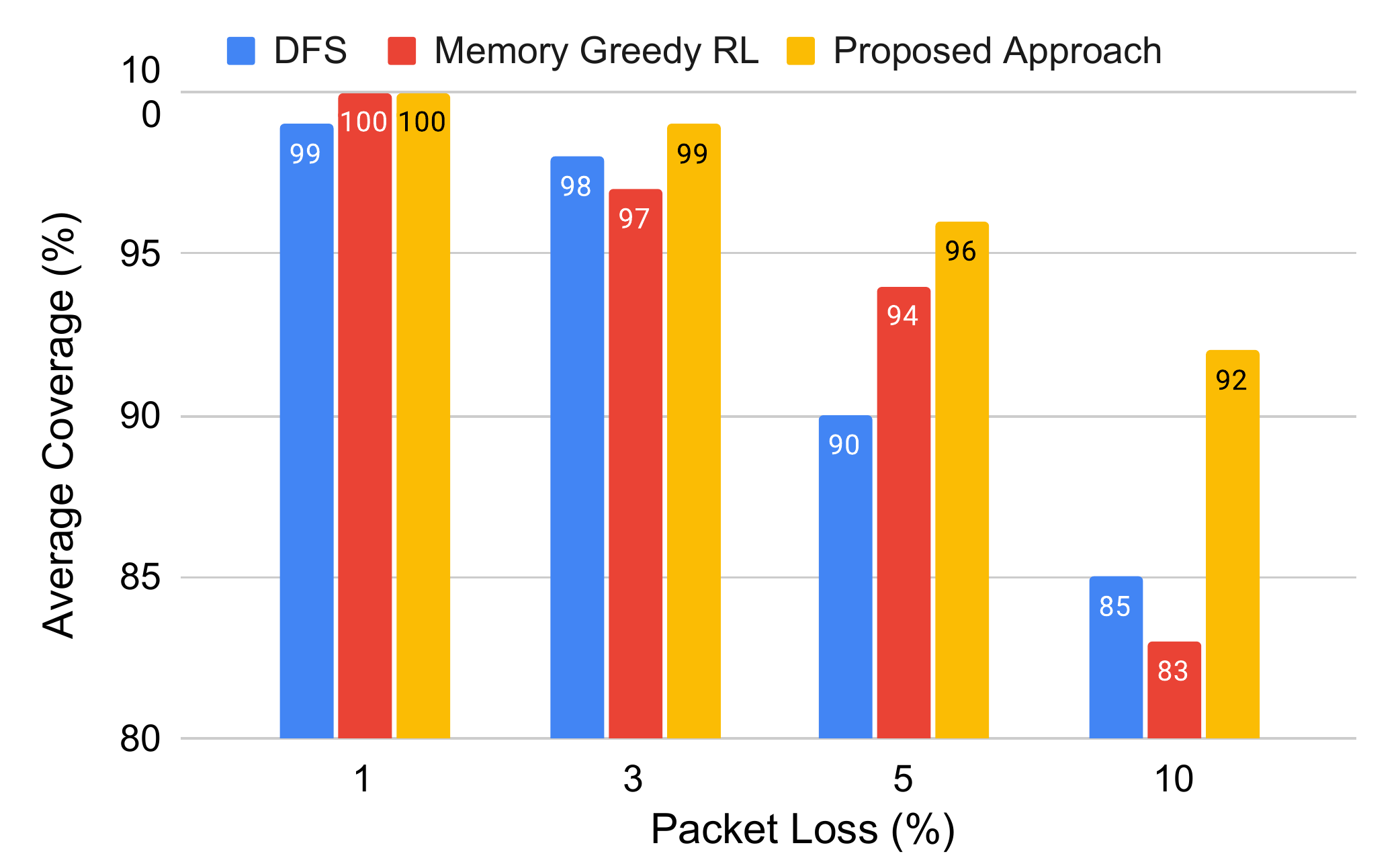}
\includegraphics[width=0.31\linewidth]{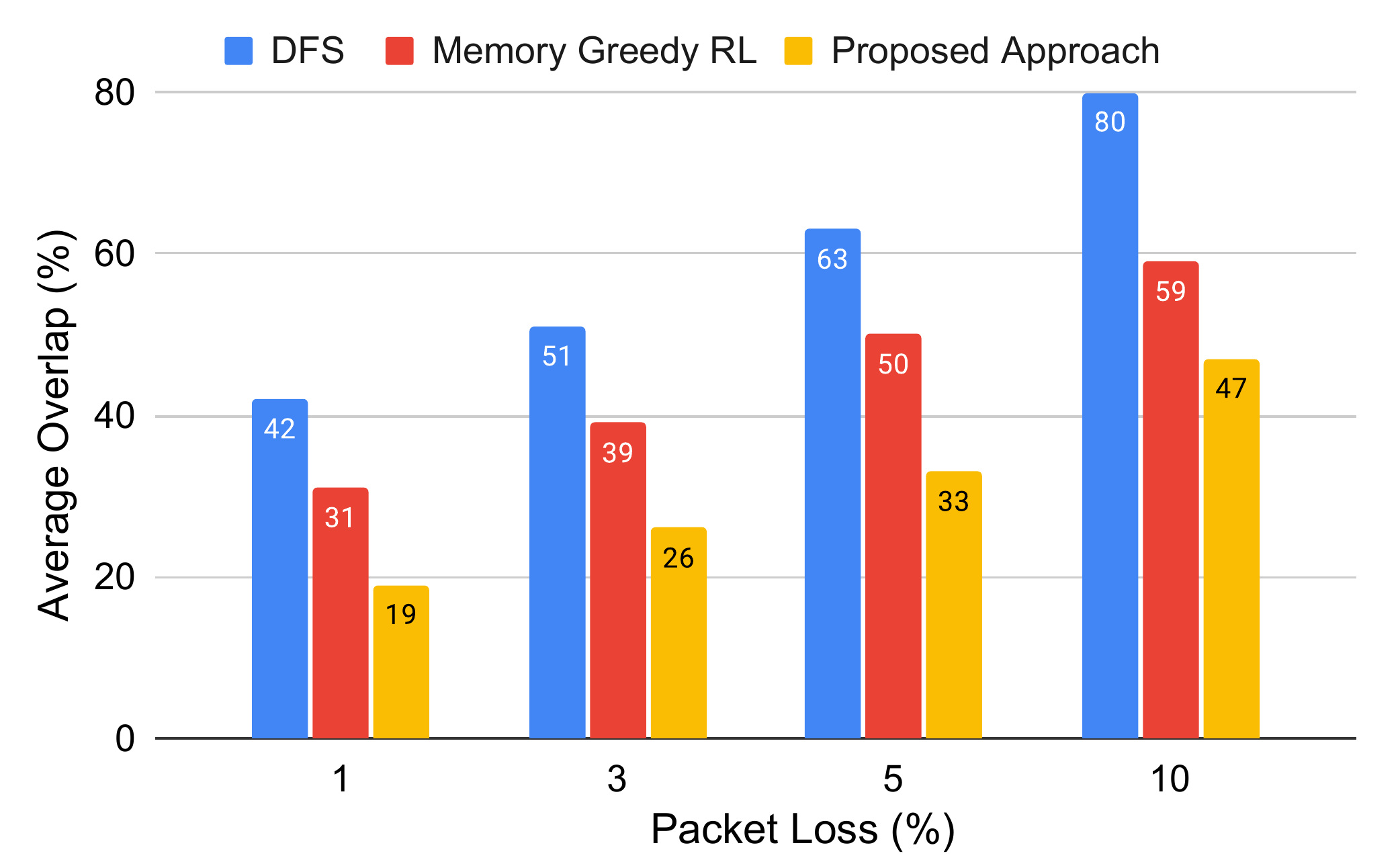}
\vspace{-4mm}
\caption{Impact of the packet loss parameter; number of iterations \textbf{(Left)}, average coverage percentage \textbf{(Center)}, average overlap percentage \textbf{(Right)}.}
    \label{loss_results}
    \vspace{-6mm}
\end{figure*}

\section{Experimental Evaluation}
We have performed simulation experiments using CORE\footnote{\url{http://coreemu.github.io/core/}} Networking emulator. All experiments were performed under standard CORE wireless network configurations other than the parameters mentioned below. In our experiments, we set up four robots on four corners of the maze (see Fig.~\ref{simulation}). Initial positions, motion constraints, communication channels, and maze dimensions are known for experimentation.  
\begin{itemize}
    \item Range: 1500 pixel distance between nodes
    \item Delay: 0.2 $\mu$sec
    \item Bandwidth: 54 Mbps
\end{itemize}

We validated the performance of the proposed approach in comparison with DFS \cite{chen2020improved}, and the memory-greedy RL \cite{yu2021memory} approaches. In the implemented approaches, the robots maintain synchronous communication and share only the current state and observed region with other robots in the maze. DFS does not incorporate received information other than updating the current exploration map, which increases the overlapping of exploration over iterations. 
We have performed experiments for 17 mazes available online\footnote{\url{http://www.tcp4me.com/mmr/mazes/}}, each with a 16$\times$16 grid size.

\subsection{Results}
\textbf{Coverage:} We look at the time efficiency to complete the maze with one-third (33\%), two-thirds (66\%), and full coverage (99\%); results can be seen in Fig.~\ref{time_per_coverage}. DFS outperformed other approaches for one-third coverage as its processing and propagation does not involve robot cooperation and explore the maze rapidly. However, DFS and our approach have comparable completion times for two-thirds coverage. The trend changes towards the end of coverage completion due to a lack of cooperation; DFS takes approx. 100, and 25 more iterations than the proposed and memory-greedy RL approaches. We also have recorded performance microscopically over every iteration. Standalone DFS works better for a few initial iterations for coverage, as its processing does not involve decision-making. However, after half of the maze coverage, the proposed approach can cover the region more rapidly than DFS. Overall, both approaches can cover 99\% of the explorable region within 400 iterations for all mazes.

We have analyzed the performance of all approaches for one maze for full coverage and observed performance for 100 and 400 iterations for all the experiment mazes. Results in Fig.~\ref{coverage_results} have shown that DFS and memory-greedy RL approaches have better coverage percentages than the proposed approach initially at 100 iterations; however, at 400 iterations, the proposed approach is approximately 10\% better in coverage. 

\textbf{Overlap:} We have analyzed the performance based on the inter-robot overlapping coverage space.
 Results (see Fig.~\ref{overlap_results}) have shown that DFS has a higher overlap than the proposed approach at both iteration cycles, which is 3\% and 20\% for 100 and 400 iterations, respectively. Interestingly, the memory-greedy RL approach balanced the charts by providing a faster coverage rate but at a loss of map overlap ratio.

\subsection{Ablation studies on communication efficiency}
\textbf{Communication Range:} We have varied the communication range from 1500 to 500-pixel units between nodes in the simulator to evaluate the performance of maze coverage strategies over completion time, percentage coverage, and percentage overlap. Fig.~\ref{range_results} shows that our approach has the most negligible impact of weak communication for coverage time due to preplanned policy utilization and strong cooperation among robots. Conversely, reduced range drastically affects the memory-greedy RL approach, and DFS moderately affects weak communication on maze coverage time. The coverage percentage reduces with the reduction in range for all approaches differently. DFS only able to cover 82\% of the overall maze with the 500 range, and memory-greedy RL covers a slightly larger area of 87\% maze than DFS. The proposed approach has shown the most negligible impact of range on the coverage and can cover the maximum region even with a range of 500 units. Regarding overlap percentage, robots in the proposed strategy intentionally avoid already explored regions due to the optimized reward function, resulting in less overlap than other strategies. DFS has high overlap during overall coverage and almost re-explored three-quarters of the whole maze in the slightest communication range.

\textbf{Packet Loss:} 
To evaluate the effectiveness of maze coverage strategies across completion time, percentage coverage, and percentage overlap, we increased the packet loss percentage in the CORE simulator from 1 to 10\%. According to the data, the proposed strategy has the most negligible influence on packet loss for coverage time (see Fig.~\ref{loss_results}) since it uses a preplanned policy and has excellent robot collaboration. On the other hand, a higher packet loss of 10\% significantly impacts the DFS technique and requires 300 more iterations to cover the labyrinth space. The memory-greedy RL had a minor impact on packet loss on the maze coverage time, but it could still cover the entire area in 753 iterations. Additionally, each technique's average coverage percentage decreases as the packet loss percentage rises. With a 10\% packet loss, memory-greedy RL covers 83\% of the entire maze, whereas DFS  covers a slightly larger area. The proposed method can cover the maximum region (92\%) even with a 10 percent packet loss and has the most negligible noticeable impact of range on coverage percentage. Due to the optimized reward function, robots in the proposed method purposefully avoid already investigated locations, resulting in less overlap than other strategies. 

\section{Conclusion}

This paper proposed a new coordination algorithm aided by reinforcement learning to solve swarm robotic exploration problems. We consider the problem of swarm robot maze coverage for a particular exploration assignment that reduces exploration time and expenditure while the energy and communication costs must be minimized.
To address this problem, we proposed a communication-efficient collaborative reinforcement learning method that uses only local information transfers to work in a distributed manner. 
We integrate a repelling mechanism similar to potential field-based strategies to avoid inter-robot collisions.
Theoretically, our approach has been compared to conventional search-based and heuristic methods, including recent variants of RL algorithms proposed in the literature. 
Through extensive CORE network simulations, we have demonstrated the superiority of our algorithm in terms of maze coverage performance even in low communication range and high packet loss environments.
The results had lower overall map overlap percentage and shorter maze coverage times, offering 20\% and 10\% improvements than the Depth First Search and memory-greedy RL approaches, respectively. 
The results provide a promising avenue for further investigations and extensions of the proposed approach to generalize to other swarm robotic network applications.

\bibliography{ref}

\begin{thebibliography}{10}
\providecommand{\url}[1]{#1}
\csname url@samestyle\endcsname
\providecommand{\newblock}{\relax}
\providecommand{\bibinfo}[2]{#2}
\providecommand{\BIBentrySTDinterwordspacing}{\spaceskip=0pt\relax}
\providecommand{\BIBentryALTinterwordstretchfactor}{4}
\providecommand{\BIBentryALTinterwordspacing}{\spaceskip=\fontdimen2\font plus
\BIBentryALTinterwordstretchfactor\fontdimen3\font minus
  \fontdimen4\font\relax}
\providecommand{\BIBforeignlanguage}[2]{{%
\expandafter\ifx\csname l@#1\endcsname\relax
\typeout{** WARNING: IEEEtran.bst: No hyphenation pattern has been}%
\typeout{** loaded for the language `#1'. Using the pattern for}%
\typeout{** the default language instead.}%
\else
\language=\csname l@#1\endcsname
\fi
#2}}
\providecommand{\BIBdecl}{\relax}
\BIBdecl

\bibitem{alamri2021autonomous}
S.~Alamri, S.~Alshehri, W.~Alshehri, H.~Alamri, A.~Alaklabi, and T.~Alhmiedat,
  ``Autonomous maze solving robotics: Algorithms and systems,''
  \emph{International Journal of Mechanical Engineering and Robotics Research},
  vol.~10, no.~12, 2021.

\bibitem{tjiharjadi2017optimization}
S.~Tjiharjadi, M.~Wijaya, and E.~Setiawan, ``Optimization maze robot using a*
  and flood fill algorithm,'' \emph{International Journal of Mechanical
  Engineering and Robotics Research}, vol.~6, no.~5, pp. 366--372, 2017.

\bibitem{chen2020improved}
Y.-H. Chen and C.-M. Wu, ``An improved algorithm for searching maze based on
  depth-first search,'' in \emph{2020 IEEE International Conference on Consumer
  Electronics-Taiwan (ICCE-Taiwan)}.\hskip 1em plus 0.5em minus 0.4em\relax
  IEEE, 2020, pp. 1--2.

\bibitem{youssefi2021swarm}
K.~A.-R. Youssefi and M.~Rouhani, ``Swarm intelligence based robotic search in
  unknown maze-like environments,'' \emph{Expert Systems with Applications},
  vol. 178, p. 114907, 2021.

\bibitem{bono2021swarm}
A.~Bono, G.~Fedele, and G.~Franz{\`e}, ``A swarm-based distributed model
  predictive control scheme for autonomous vehicle formations in uncertain
  environments,'' \emph{IEEE Transactions on Cybernetics}, vol.~52, no.~9, pp.
  8876--8886, 2021.

\bibitem{kalinowska2022over}
A.~Kalinowska, E.~Davoodi, F.~Strub, K.~W. Mathewson, I.~Kajic, M.~Bowling,
  T.~D. Murphey, and P.~M. Pilarski, ``Over-communicate no more: Situated rl
  agents learn concise communication protocols,'' \emph{arXiv preprint
  arXiv:2211.01480}, 2022.

\bibitem{gu2021improved}
S.~Gu and G.~Mao, ``An improved q-learning algorithm for path planning in maze
  environments,'' in \emph{Intelligent Systems and Applications: Proceedings of
  the 2020 Intelligent Systems Conference (IntelliSys) Volume 2}.\hskip 1em
  plus 0.5em minus 0.4em\relax Springer, 2021, pp. 547--557.

\bibitem{shi2020efficient}
L.~Shi, S.~Li, Q.~Zheng, M.~Yao, and G.~Pan, ``Efficient novelty search through
  deep reinforcement learning,'' \emph{IEEE Access}, vol.~8, pp.
  128\,809--128\,818, 2020.

\bibitem{yu2021memory}
X.~Yu, Y.~Wu, X.-M. Sun, and W.~Zhou, ``A memory-greedy policy with guaranteed
  convergence for accelerating reinforcement learning,'' \emph{Journal of
  Autonomous Vehicles and Systems}, vol.~1, no.~1, p. 011005, 2021.

\bibitem{tjiharjadi2022systematic}
S.~Tjiharjadi, S.~Razali, and H.~A. Sulaiman, ``A systematic literature review
  of multi-agent pathfinding for maze research,'' \emph{Journal of Advances in
  Information Technology Vol}, vol.~13, no.~4, 2022.

\bibitem{luis2020using}
N.~Luis, T.~Pereira, S.~Fern{\'a}ndez, A.~Moreira, D.~Borrajo, and M.~Veloso,
  ``Using pre-computed knowledge for goal allocation in multi-agent planning,''
  \emph{Journal of Intelligent \& Robotic Systems}, vol.~98, no.~1, pp.
  165--190, 2020.

\bibitem{shi2019multi}
W.~Shi, J.~Li, N.~Cheng, F.~Lyu, S.~Zhang, H.~Zhou, and X.~Shen, ``Multi-drone
  3-d trajectory planning and scheduling in drone-assisted radio access
  networks,'' \emph{IEEE Transactions on Vehicular Technology}, vol.~68, no.~8,
  pp. 8145--8158, 2019.

\bibitem{wang2019coordination}
L.~Wang and Q.~Guo, ``Coordination of multiple autonomous agents using
  naturally generated languages in task planning,'' \emph{Applied Sciences},
  vol.~9, no.~17, p. 3571, 2019.

\bibitem{pozza2021quantum}
N.~D. Pozza, L.~Buffoni, S.~Martina, and F.~Caruso, ``Quantum reinforcement
  learning: the maze problem,'' \emph{arXiv preprint arXiv:2108.04490}, 2021.

\bibitem{yang2021can}
Q.~Yang and R.~Parasuraman, ``How can robots trust each other for better
  cooperation? a relative needs entropy based robot-robot trust assessment
  model,'' in \emph{2021 IEEE International Conference on Systems, Man, and
  Cybernetics (SMC)}.\hskip 1em plus 0.5em minus 0.4em\relax IEEE, 2021, pp.
  2656--2663.

\bibitem{kantasewi2019multi}
N.~Kantasewi, S.~Marukatat, S.~Thainimit, and O.~Manabu, ``Multi q-table
  q-learning,'' in \emph{2019 10th International Conference of Information and
  Communication Technology for Embedded Systems (IC-ICTES)}.\hskip 1em plus
  0.5em minus 0.4em\relax IEEE, 2019, pp. 1--7.

\bibitem{uwano2017communication}
F.~Uwano and K.~Takadama, ``Communication-less cooperative q-learning agents in
  maze problem,'' in \emph{Intelligent and Evolutionary Systems}.\hskip 1em
  plus 0.5em minus 0.4em\relax Springer, 2017, pp. 453--467.

\bibitem{yu2019navigation}
X.~Yu, Y.~Wu, and X.-M. Sun, ``A navigation scheme for a random maze using
  reinforcement learning with quadrotor vision,'' in \emph{2019 18th European
  Control Conference (ECC)}.\hskip 1em plus 0.5em minus 0.4em\relax IEEE, 2019,
  pp. 518--523.

\bibitem{latif2022dgorl}
E.~Latif and R.~Parasuraman, ``Dgorl: Distributed graph optimization based
  relative localization of multi-robot systems,'' \emph{arXiv preprint
  arXiv:2210.01662}, 2022.

\bibitem{viseras2016planning}
A.~Viseras, R.~O. Losada, and L.~Merino, ``Planning with ants: Efficient path
  planning with rapidly exploring random trees and ant colony optimization,''
  \emph{International Journal of Advanced Robotic Systems}, vol.~13, no.~5, p.
  1729881416664078, 2016.

\bibitem{ozdemir2019spatial}
A.~{\"O}zdemir, M.~Gauci, A.~Kolling, M.~D. Hall, and R.~Gro{\ss}, ``Spatial
  coverage without computation,'' in \emph{2019 International Conference on
  Robotics and Automation (ICRA)}.\hskip 1em plus 0.5em minus 0.4em\relax IEEE,
  2019, pp. 9674--9680.

\bibitem{surynek2016efficient}
P.~Surynek, A.~Felner, R.~Stern, and E.~Boyarski, ``Efficient sat approach to
  multi-agent path finding under the sum of costs objective,'' in
  \emph{Proceedings of the twenty-second european conference on artificial
  intelligence}, 2016, pp. 810--818.

\end{thebibliography}
\bibliographystyle{IEEEtran}

\end{document}